\PassOptionsToPackage{bookmarks=false}{hyperref}
\documentclass[cameraready]{Interspeech}

\usepackage{booktabs}
\usepackage{multirow}
\usepackage{amsmath,amssymb}
\usepackage{graphicx}
\usepackage{tikz}
\usepackage{pifont}
\usetikzlibrary{positioning, arrows.meta, fit, backgrounds, calc, decorations.pathreplacing}

\definecolor{cblue}{HTML}{1D4ED8}     
\definecolor{corange}{HTML}{C2410C}   
\definecolor{cred}{HTML}{B91C1C}      
\definecolor{cpurple}{HTML}{7E22CE}   
\definecolor{cteal}{HTML}{0F766E}     
\definecolor{cgreen}{HTML}{15803D}    
\definecolor{csty}{HTML}{0E7490}      
\definecolor{ccon}{HTML}{B45309}      
\definecolor{cfun}{HTML}{64748B}      

\let\tipaBang\!
\DeclareRobustCommand{\!}{\ifmmode\mskip-\thinmuskip\else\tipaBang\fi}
\let\tipaSemi\;
\DeclareRobustCommand{\;}{\ifmmode\mskip\thickmuskip\else\tipaSemi\fi}

\title{How Do Instructions Shape Speech? Cross-Attention Attribution\\for Style-Captioned Text-to-Speech}

\author[affiliation={1}]{Nityanand}{Mathur}
\author[affiliation={1}]{Hamees}{Sayed}
\author[affiliation={1}]{Wasim}{Madha}
\author[affiliation={1}]{Apoorv}{Singh}
\author[affiliation={1}]{Sameer}{Khurana}
\author[affiliation={1}]{Akshat}{Mandloi}
\author[affiliation={1}]{Sudarshan}{Kamath}
\address{
    $^1$ Smallest.ai
}
\email{nityanandmathur@gmail.com}

\begin{document}

\maketitle

\begin{abstract}
Style-captioned text-to-speech systems use natural language to control voice characteristics, but how individual words influence acoustic output remains unclear. Understanding this is critical for diagnosing failure modes and improving controllability in expressive TTS. We propose cross-attention attribution for speech diffusion models, adapting the DAAM framework to the speech domain for the first time, and apply it to CapSpeech-TTS. Our method extracts per-token heatmaps across 25 layers and 24 ODE steps. We analyze 3{,}600 (style caption, text transcript) combinations comprising 120 style captions conditioning the generation of 30 text transcripts each, revealing how caption tokens shape waveforms. Results show: (1) style tokens have lower temporal variance than content/function tokens, confirming global conditioning; (2) style attention correlates with F0 and energy; (3) style conditioning peaks in early steps and deep layers; (4) attention entropy reaches its minimum at layer 17, co-occurring with the style importance peak, indicating maximal network selectivity at the most style-critical stage. This is the first study of how natural language influences cross-attention in speech diffusion models.
\end{abstract}

\section{Introduction}
\label{sec:intro}

Modern text-to-speech (TTS) systems have moved beyond fixed speaker embeddings~\cite{dehak2011front} toward natural language conditioning, where free-form captions such as ``a calm, deep voice speaking slowly'' control the style of generated speech~\cite{wang2024capspeech, le2024voicebox, tan2024naturalspeech3}. This paradigm, building on advances in generative spoken language modeling~\cite{lakhotia2021generative}, offers powerful expressivity but introduces a fundamental interpretability question: \emph{how do individual words in a style caption influence the synthesized audio?}

In text-to-image generation, Diffusion Attentive Attribution Maps (DAAM)~\cite{tang2023daam} answered an analogous question by extracting cross-attention maps from Stable Diffusion~\cite{rombach2022ldm} and attributing spatial regions to prompt tokens. Prompt-to-Prompt~\cite{hertz2023p2p} and generic attention rollout~\cite{chefer2021generic} further established cross-attention as a faithful proxy for conditioning influence. No equivalent framework exists for speech. Unlike images, speech is inherently temporal: attention must map caption tokens to spectrogram time frames rather than spatial pixels. Furthermore, the distinction between \emph{global} properties (style, emotion) and \emph{local} properties (phoneme identity, duration) creates an interpretability challenge unique to speech.

We address this gap by adapting DAAM to CapSpeech~\cite{wang2024capspeech}, a non-autoregressive TTS model that uses flow matching~\cite{lipman2023flow} conditioned on T5-encoded~\cite{raffel2020t5} style captions. Our contributions are:
\begin{itemize}[leftmargin=1.5em]
    \item[\ding{182}] The \textbf{first cross-attention attribution analysis for TTS}, extracting per-token temporal heatmaps across all 25 transformer layers and 24 ODE steps over 3{,}600 (style caption, text transcript) combinations.
    \item[\ding{183}] A \textbf{global-vs-local analysis} showing that style tokens exhibit significantly lower temporal variance ($p < 10^{-43}$, $d = {-}1.16$) than content and function tokens.
    \item[\ding{184}] An \textbf{acoustic grounding analysis} demonstrating that style-token attention correlates with F0 and energy with semantic coherence (e.g., ``loud'' $\leftrightarrow$ energy, $r = +0.64$).
    \item[\ding{185}] A \textbf{layer-step dynamics study} revealing hierarchical conditioning: style importance peaks in early ODE steps ($5.2\times$ decay) and deepens through transformer layers.
\end{itemize}

\section{Related Work}
\label{sec:related}

\textbf{Style-conditioned TTS.} Recent TTS systems accept natural language descriptions to control voice style. CapSpeech~\cite{wang2024capspeech} conditions a flow-matching transformer on T5-encoded captions, while VoiceBox~\cite{le2024voicebox} and NaturalSpeech~3~\cite{tan2024naturalspeech3} explore similar paradigms. Earlier autoregressive approaches like Tacotron~\cite{wang2017tacotron}, Tacotron~2~\cite{shen2018natural}, and FastSpeech~\cite{ren2020fastspeech} established attention-based TTS but relied on fixed speaker embeddings~\cite{dehak2011front} rather than flexible text conditioning. Zero-shot multi-speaker systems like YourTTS~\cite{casanova2022yourtts} and NaturalSpeech~2~\cite{yuan2024naturalspeech2} further demonstrated the potential of latent diffusion for voice control. Despite impressive generation quality, the internal mechanisms by which these systems process style instructions remain unexplored.

\textbf{Interpretability in generative models.} DAAM~\cite{tang2023daam} showed that cross-attention maps in Stable Diffusion~\cite{rombach2022ldm} faithfully attribute image regions to individual prompt tokens. Prompt-to-Prompt~\cite{hertz2023p2p} leveraged these maps for controlled image editing, while Chefer et al.~\cite{chefer2021generic} proposed generic attention rollout for encoder-decoder transformers. In speech, attention analysis has been limited to alignment visualization in autoregressive models~\cite{ren2021fastspeech2}. Our work brings DAAM-style attribution to the speech domain for the first time.

\textbf{Flow matching for speech.} Flow matching~\cite{lipman2023flow} offers a simulation-free alternative to diffusion probabilistic models~\cite{ho2020ddpm,song2021denoising} by directly learning the velocity field of the probability flow ODE. Unlike diffusion models that require iterative denoising through a learned score function, flow matching provides a deterministic mapping from noise to data via continuous normalizing flows. This formulation has gained traction in speech synthesis, with models like Grad-TTS~\cite{popov2021gradtts} pioneering diffusion-based mel-spectrogram generation, Glow-TTS~\cite{kim2021conditional} using normalizing flows, and F5-TTS~\cite{chen2024f5tts} demonstrating flow matching for high-quality synthesis. CapSpeech adopts flow matching with a Diffusion Transformer (DiT)~\cite{peebles2023dit} backbone, processing mel-spectrogram latents through a transformer~\cite{vaswani2017attention} with explicit cross-attention at every layer and ODE step. This design makes it particularly amenable to DAAM-style attribution: unlike autoregressive models where attention primarily serves alignment, the DiT's cross-attention directly conditions the velocity field on caption embeddings at each generation stage, providing a clear interpretable pathway from text to acoustics.

\section{Method}
\label{sec:method}

\subsection{CapSpeech architecture overview}





\begin{figure}[t]
\centering
\resizebox{\columnwidth}{!}{%
\begin{tikzpicture}[
    node distance=0.6cm and 0.8cm,
    box/.style={draw, rounded corners=3pt, minimum height=0.8cm, minimum width=1.8cm, font=\footnotesize, align=center, fill=#1},
    box/.default=blue!8,
    arrow/.style={-{Stealth[length=2.5mm]}, thick},
    darrow/.style={-{Stealth[length=2.5mm]}, thick, dashed, red!70!black},
]

\node[box=blue!15] (selfattn) {Self-Attn};
\node[box=blue!30, right=0.4cm of selfattn] (crossattn) {Cross-Attn};
\node[box=blue!15, right=0.4cm of crossattn] (ffn) {FFN};

\node[box=orange!30, above=1cm of crossattn] (t5) {T5 Encoder};
\node[box=orange!15, left=0.6cm of t5] (caption) {Style Caption};

\begin{scope}[on background layer]
\node[draw, rounded corners=5pt, fill=blue!8,
      fit=(selfattn)(crossattn)(ffn),
      inner sep=0.25cm] (dit) {};
\end{scope}

\draw[arrow, thin] (selfattn) -- (crossattn);
\draw[arrow, thin] (crossattn) -- (ffn);

\node[minimum width=1.4cm, minimum height=1.0cm, left=0.8cm of dit] (noise) {};
\draw[gray!60, semithick] plot[smooth, tension=0.6] coordinates {
    ([xshift=-0.45cm, yshift=-0.25cm]noise.center)
    ([xshift=-0.3cm, yshift=-0.2cm]noise.center)
    ([xshift=-0.15cm, yshift=0.0cm]noise.center)
    ([xshift=-0.05cm, yshift=0.2cm]noise.center)
    ([yshift=0.3cm]noise.center)
    ([xshift=0.05cm, yshift=0.2cm]noise.center)
    ([xshift=0.15cm, yshift=0.0cm]noise.center)
    ([xshift=0.3cm, yshift=-0.2cm]noise.center)
    ([xshift=0.45cm, yshift=-0.25cm]noise.center)
};
\node[font=\scriptsize, text=gray!70] at ([yshift=-0.4cm]noise.center) {$\mathcal{N}(0,I)$};

\node[box=purple!20, right=0.8cm of dit] (vocoder) {Vocoder};

\node[minimum width=1.4cm, minimum height=1.0cm, right=0.8cm of vocoder] (wav) {};
\draw[purple!60, semithick] plot[smooth, tension=0.7] coordinates {
    ([xshift=-0.45cm]wav.center) ([xshift=-0.3cm, yshift=0.2cm]wav.center)
    ([xshift=-0.15cm, yshift=-0.05cm]wav.center) ([xshift=-0.05cm, yshift=0.28cm]wav.center)
    (wav.center) ([xshift=0.05cm, yshift=-0.26cm]wav.center)
    ([xshift=0.15cm, yshift=0.05cm]wav.center) ([xshift=0.3cm, yshift=-0.15cm]wav.center)
    ([xshift=0.45cm]wav.center)
};
\node[font=\scriptsize, text=purple!50] at ([yshift=-0.4cm]wav.center) {Waveform};

\node[box=red!15, below=1.0cm of crossattn, minimum width=2.0cm] (daam) {\textbf{DAAM Hook}};
\node[box=red!25, right=0.6cm of daam, minimum width=2.0cm] (heatmap) {Heatmaps $M_j$};

\draw[arrow] (noise) -- (dit);
\draw[arrow] (dit) -- (vocoder);
\draw[arrow] (vocoder) -- (wav);

\draw[arrow] (caption) -- (t5);
\draw[arrow] (t5) -- (crossattn);

\draw[darrow] (crossattn) -- (daam) node[midway, right, font=\scriptsize, text=red!60!black] {hook};
\draw[darrow] (daam) -- (heatmap);

\end{tikzpicture}%
}
\caption{CapSpeech pipeline with our {\color{cred}DAAM attribution hook}. {\color{corange}Style captions} condition the {\color{cblue}Flow-Matching DiT}. Cross-attention maps are intercepted at every layer and ODE step ({\color{cred}dashed arrows}), then aggregated into per-token temporal heatmaps ${\color{cred}M_j}$.}
\label{fig:architecture}
\end{figure}
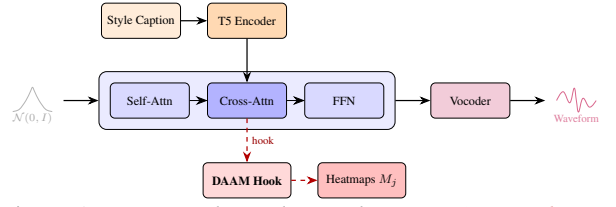

Figure~\ref{fig:architecture} illustrates the CapSpeech pipeline and our DAAM attribution mechanism. The model comprises four components working in sequence to transform a style caption and text transcript into a waveform. The pipeline begins with a \textbf{T5 caption encoder}~\cite{raffel2020t5}, which maps the style caption $\mathbf{c} = (c_1, \ldots, c_{T_c})$ to contextual embeddings $\mathbf{E}_c \in \mathbb{R}^{T_c \times d}$; T5's pre-trained encoder provides rich semantic representations that capture linguistic nuances in style descriptions. These text embeddings are complemented by a \textbf{CLAP encoder}~\cite{wu2023clap}, which produces a global style tag embedding $\mathbf{e}_\text{clap} \in \mathbb{R}^{d'}$ from a short style label, providing acoustic grounding alongside the T5 representations. The central component is a \textbf{flow-matching DiT}~\cite{lipman2023flow, chen2024f5tts}, which serves as the core generative model: a transformer with $L{=}25$ layers that iteratively refines a mel-spectrogram latent $\mathbf{x}_s \in \mathbb{R}^{T_a \times d_\text{mel}}$ from Gaussian noise $\mathbf{x}_0 \sim \mathcal{N}(0, I)$ over $S{=}24$ ODE steps. Each layer of the DiT contains self-attention, cross-attention, and feed-forward sublayers, where the cross-attention mechanism is the locus of style conditioning: queries from the current latent $\mathbf{x}_s$ attend to keys and values derived from the caption embeddings $\mathbf{E}_c$. Finally, a \textbf{HiFi-GAN vocoder}~\cite{kong2020hifigan} converts the refined mel-spectrogram into the output waveform $\mathbf{w} \in \mathbb{R}^{N}$.

\noindent In flow matching, the model learns a velocity field $v_\theta(\mathbf{x}_s, s, \mathbf{E}_c)$ satisfying the ODE:
\begin{equation}
    \frac{d\mathbf{x}}{ds} = v_\theta\!\bigl(\mathbf{x}_s,\; s,\; \mathbf{E}_c\bigr), \quad s \in [0, 1]
    \label{eq:ode}
\end{equation}
transporting noise $\mathbf{x}_0 \sim \mathcal{N}(0,I)$ to the data distribution. At each ODE step $s$ and each layer $l$, the transformer computes multi-head cross-attention:
\begin{equation}
    A^{(l,s)}_h = \mathrm{softmax}\!\left(\frac{\mathbf{Q}^{(l,s)}_h\; {\mathbf{K}^{(l,s)}_h}^\top}{\sqrt{d_k}}\right) \in \mathbb{R}^{T_a \times T_c}
    \label{eq:attention}
\end{equation}
where $h \in \{1, \ldots, H\}$ indexes attention heads, and $\mathbf{Q}^{(l,s)} \in \mathbb{R}^{T_a \times d_k}$ is the audio query, $\mathbf{K}^{(l,s)} \in \mathbb{R}^{T_c \times d_k}$ is the caption key. This cross-attention is the mechanism through which the caption \emph{directly} influences audio generation at every layer and every step.

\subsection{DAAM adaptation for speech}
\label{sec:daam}

We register forward hooks on each cross-attention module to intercept $A^{(l,s)}_h$ (Eq.~\ref{eq:attention}) at every layer $l \in \{0, \ldots, L{-}1\}$ and ODE step $s \in \{0, \ldots, S{-}1\}$. For each tensor $A^{(l,s)} \in \mathbb{R}^{H \times T_a \times T_c}$, we first average over heads:
\begin{equation}
    \bar{A}^{(l,s)} = \frac{1}{H} \sum_{h=1}^{H} A^{(l,s)}_h \in \mathbb{R}^{T_a \times T_c}
    \label{eq:head_avg}
\end{equation}

We then aggregate across all layers and steps to obtain a per-token heatmap:
\begin{equation}
    M_j = \frac{1}{L \cdot S} \sum_{l=0}^{L-1} \sum_{s=0}^{S-1} \bar{A}^{(l,s)}_{:,\, j} \in \mathbb{R}^{T_a}
    \label{eq:aggregate}
\end{equation}
where $M_j$ is the temporal attribution map for caption token $j$. This 1-D heatmap is analogous to the 2-D spatial maps in image DAAM~\cite{tang2023daam}, but over the audio time axis. In total, we capture $L \times S = 600$ attention matrices per generation.

\subsection{Token categorisation}
\label{sec:token_cat}

We classify each caption token into three functional categories:

\begin{itemize}
    \item \textbf{{\color{csty}Style}} ($\mathcal{C}_\text{sty}$): 30~adjectives describing voice quality, emotion, or pace (e.g., ``calm'', ``bright'', ``harsh''). 7{,}968~instances across all generations.
    \item \textbf{{\color{ccon}Content}} ($\mathcal{C}_\text{con}$): 20~nouns describing the speaker or voice (e.g., ``voice'', ``speaker'', ``male''). 8{,}480~instances across all generations.
    \item \textbf{{\color{cfun}Function}} ($\mathcal{C}_\text{fn}$): articles, prepositions, punctuation. 38{,}432~instances across all generations.
\end{itemize}

\noindent The style adjectives and content nouns were curated from TTS literature~\cite{wang2024capspeech,le2024voicebox}, covering emotional, prosodic, and quality attributes. Since T5's SentencePiece tokenizer splits words into subword units, we merge contiguous subwords by detecting word-boundary prefixes before classification.

\subsection{Analysis metrics}
\label{sec:metrics}

We define five complementary metrics on each token's aggregated heatmap $M_j$.

\textbf{Temporal variance} $\sigma^2_j = \frac{1}{T_a}\sum_t (M_{j,t} - \bar{M}_j)^2$ measures how concentrated attention is over time. Low variance indicates temporally diffuse (global) influence; high variance signals time-localized attention. \textbf{Peak-to-mean ratio} $\text{PMR}_j = \max_t M_{j,t}\, / \,(\bar{M}_j + 10^{-8})$ captures the sharpness of the attention peak relative to the background level. A ratio near 1 indicates uniform attention; values $\gg 1$ indicate a dominant spike at a particular time region. \textbf{Temporal entropy} $H_j = -\sum_t p_{j,t} \log_2 p_{j,t}$ (where $p_{j,t} = M_{j,t} / \sum_{t'} M_{j,t'}$) quantifies distributional uniformity in the information-theoretic sense~\cite{shannon1949communication}. Maximum entropy $H_\text{max} = \log_2 T_a$ corresponds to perfectly uniform attention across all audio frames.

\textbf{Acoustic correlation.} We extract frame-level F0 (pYIN algorithm~\cite{maas2011pyin}, 50--600\,Hz range to cover typical male and female speech) and RMS energy from each generated waveform using librosa~\cite{mcfee2015librosa}. Both features are linearly interpolated to $T_a$ frames to match the attention time axis, and we compute Pearson correlation $r$ between each token's heatmap $M_j$ and each acoustic feature $f \in \{\text{F0}, \text{energy}\}$. A positive correlation indicates that the model allocates more attention to a given token in temporal regions where that acoustic feature is elevated.

\textbf{Layer/step importance.} To understand where in the generation process each token category is most influential, we compute per-category importance $I^{(l)}_\mathcal{C}$ at each layer~$l$ by averaging, for all tokens $j$ in category $\mathcal{C}$, the mean attention weight received across all ODE steps and audio frames:
\begin{equation}
    I^{(l)}_\mathcal{C} = \frac{1}{S \cdot T_a \cdot |\mathcal{C}|} \sum_{s=0}^{S-1} \sum_{t=1}^{T_a} \sum_{j \in \mathcal{C}} \bar{A}^{(l,s)}_{t,j}
    \label{eq:layer_importance}
\end{equation}
Step importance $I^{(s)}_\mathcal{C}$ is defined analogously by averaging across layers. We further summarise these as a late-to-early ratio $R_\mathcal{C}$, comparing mean importance in layers 13--24 to layers 0--12, and a step decay ratio $D_\mathcal{C} = I^{(0)}_\mathcal{C} / I^{(S-1)}_\mathcal{C}$.

\subsection{Dataset}
\label{sec:dataset}

We construct 120 style captions by systematically combining 30 style adjectives (see token vocabulary after references) with 20 content nouns across 6 caption templates (e.g., ``a \{adjective\} \{noun\} speaking''). Each style caption serves as the cross-attention conditioning input to the model. We pair these with 30 diverse text transcripts (varying in length, phonetic complexity, and content) for which audio is generated. This yields $120 \times 30 = 3{,}600$ \emph{(style caption, text transcript)} combinations. All combinations are processed through CapSpeech, yielding 3{,}520 successful generations (80 excluded due to duration estimation failures). Across 3{,}520 generations, we analyze 54{,}880 token instances and capture $3{,}520 \times 600 = 2{,}112{,}000$ attention matrices.

\section{Analysis and Results}
\label{sec:results}

\subsection{Experiment 1: Global vs.\ local conditioning}
\label{sec:fig1}

We compute temporal variance, PMR, and entropy for each token's aggregated heatmap and group by category. To quantify effect magnitude beyond $p$-values, we report Cohen's~$d$~\cite{cohen1988statistical} (standardized mean difference) alongside Mann--Whitney $U$ p-values~\cite{mann1947test}: the former provides interpretable effect sizes, while the latter avoids parametric assumptions about the underlying distributions.

\begin{figure}[h]
  \centering
  \includegraphics[width=\linewidth]{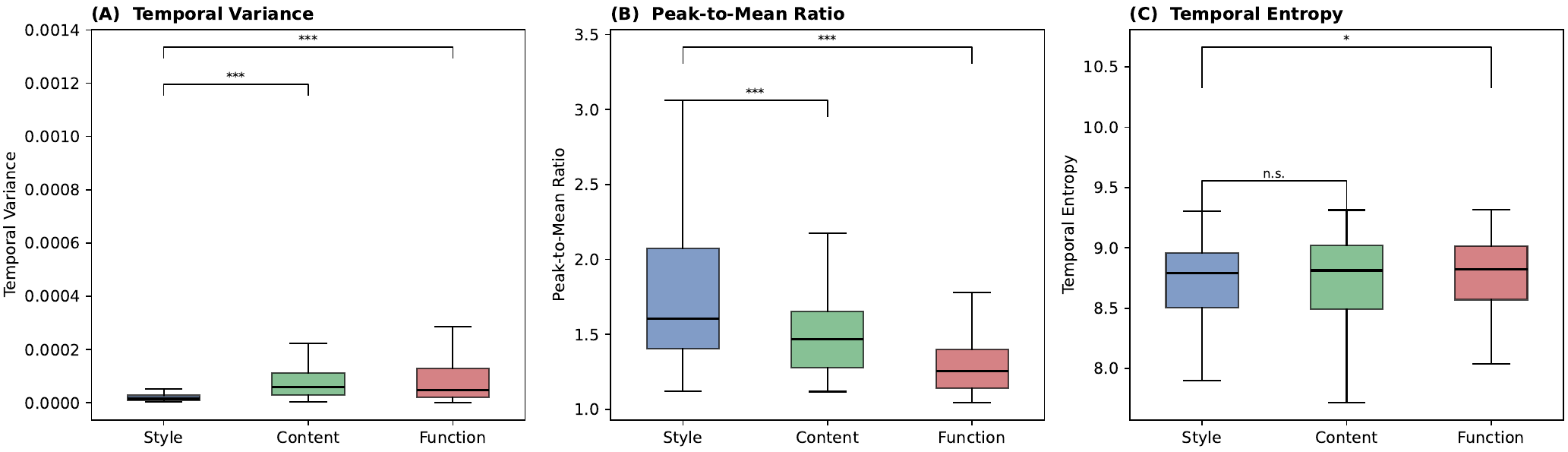}
  \caption{Global vs.\ local conditioning. Boxplots of (A)~temporal variance, (B)~PMR, and (C)~temporal entropy by category: {\color{csty}\textbf{Style}}, {\color{ccon}\textbf{Content}}, {\color{cfun}\textbf{Function}}. Brackets: Mann--Whitney $U$; {***}\,$p\!<\!0.001$.}
  \label{fig:figure1}
\end{figure}

Results are shown in Figure~\ref{fig:figure1} and Table~\ref{tab:fig1_stats}. Style tokens exhibit the \emph{lowest} temporal variance ($\bar{\sigma}^2 = 2.1 \times 10^{-5}$), significantly lower than content tokens ($7.0 \times 10^{-5}$; $p < 10^{-43}$, $d = {-}1.16$) and function tokens ($19.2 \times 10^{-5}$; $p < 10^{-44}$, $d = {-}0.72$). The $9.2\times$ variance ratio between function and style tokens confirms that style adjectives distribute attention \emph{uniformly} across the utterance, acting as global modulators rather than aligning to specific temporal regions.

Conversely, style tokens show the \emph{highest} PMR ($1.74$ vs.\ $1.48$ for content, $1.36$ for function; all $p < 10^{-10}$). This seemingly paradoxical combination, low variance yet high PMR, reveals that style tokens produce \emph{compact, characteristic attention signatures}: a consistent shape with a distinctive peak, spread uniformly across time. Function tokens show the opposite pattern (high variance, low PMR). Temporal entropy shows no significant inter-category differences ($p > 0.05$).

\textbf{Per-word variance analysis.} Table~\ref{tab:perword_var} reports temporal variance for individual style words, revealing a clear hierarchy. Words describing global prosodic properties ``cheerful'' ($\sigma^2 = 1.0 \times 10^{-5}$), ``deep'' ($1.1 \times 10^{-5}$), ``harsh'' ($1.1 \times 10^{-5}$) show the most temporally diffuse attention, while words with more salient acoustic correlates ``loud'' ($6.3 \times 10^{-5}$), ``nasal'' ($4.2 \times 10^{-5}$) exhibit higher variance, suggesting they partially modulate specific temporal regions where their acoustic effect is strongest. Even the least global style word (``loud'') still has $3.1\times$ lower variance than the function-token average.

\vspace{-6mm}
\begin{table}[h]
  \caption{Token-level metrics (mean $\pm$ std). All significance tests: Mann--Whitney~$U$. $d$: Cohen's effect size vs.\ style.}
  \label{tab:fig1_stats}
  \centering
  \small
  \resizebox{\columnwidth}{!}{%
  \begin{tabular}{@{}lccccr@{}}
    \toprule
    \textbf{Cat.} & $n$ & $\bar{\sigma}^2$~($\!\times\!10^{-5}$) & \textbf{PMR} & $\bar{H}$~(bits) & $d_\text{var}$ \\
    \midrule
    {\color{csty}Style}    &  7{,}968 & $2.1 \pm 2.2$ & $1.74 \pm 0.48$ & $8.72 \pm 0.36$ & --- \\
    {\color{ccon}Content}  &  8{,}480 & $7.0 \pm 5.6$ & $1.48 \pm 0.30$ & $8.74 \pm 0.36$ & $-1.16$ \\
    {\color{cfun}Function} & 38{,}432 & $19.2 \pm 33.5$ & $1.36 \pm 0.43$ & $8.76 \pm 0.36$ & $-0.72$ \\
    \bottomrule
  \end{tabular}%
  }
\end{table}
\vspace{-6mm}
\begin{table}[h]
  \caption{Per-word temporal variance ($\times 10^{-5}$) for selected style words, sorted by $\sigma^2$. Lower values indicate more global attention.}
  \label{tab:perword_var}
  \centering
  \small
  \begin{tabular}{@{}lrc|lrc@{}}
    \toprule
    \textbf{Word} & $n$ & $\bar{\sigma}^2$ & \textbf{Word} & $n$ & $\bar{\sigma}^2$ \\
    \midrule
    cheerful  & 640 & 1.0 & nervous   & 352 & 2.2 \\
    deep      & 320 & 1.1 & calm      & 224 & 2.4 \\
    harsh     & 320 & 1.1 & robotic   & 384 & 2.7 \\
    soft      & 320 & 1.3 & clear     & 416 & 3.7 \\
    cold      & 448 & 1.3 & dramatic  & 544 & 3.7 \\
    smooth    & 416 & 1.4 & nasal     & 288 & 4.2 \\
    excited   & 384 & 1.4 & loud      & 256 & 6.3 \\
    \bottomrule
  \end{tabular}
\end{table}

\subsection{Experiment 2: Acoustic feature correlations}
\label{sec:fig2}

We compute Pearson correlations between each token's attention heatmap $M_j$ and the F0 and energy contours of the generated waveform to test whether attention reflects measurable acoustic influence.

\begin{figure}[h]
  \centering
  \includegraphics[width=\linewidth]{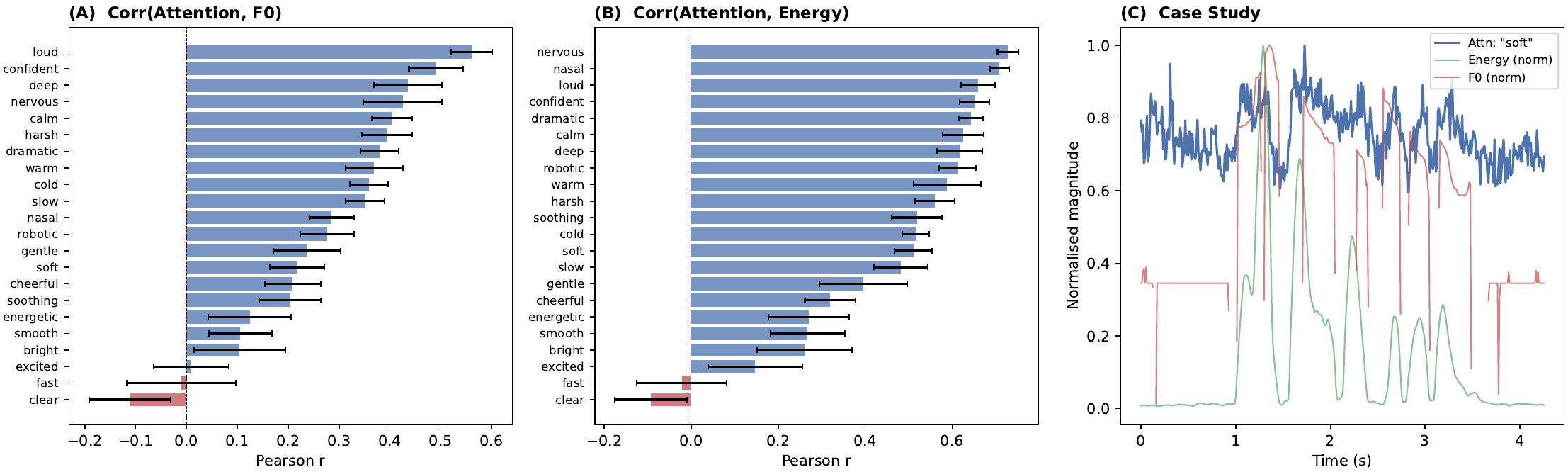}
  \caption{Acoustic grounding. (A)~Mean $r$({\color{cblue}attention}, {\color{cgreen}F0}) per style word. (B)~Same for {\color{cgreen}energy}. (C)~Case study: normalised {\color{cblue}attention}, {\color{cgreen}F0}, and {\color{cgreen}energy} overlaid.}
  \label{fig:figure2}
\end{figure}

Figure~\ref{fig:figure2} and Table~\ref{tab:fig2_corr} show style tokens have moderate positive correlations with F0 ($\bar{r} = +0.21$) and energy ($\bar{r} = +0.28$), substantially stronger than function tokens (F0: $+0.11$, energy: $+0.09$). The energy difference is highly significant ($p < 10^{-8}$); the F0 difference is also significant ($p = 0.02$). Content tokens show the strongest correlations overall (F0: $+0.50$, energy: $+0.54$), because speaker-identity words (``male'', ``female'') impose broad, categorical constraints on the pitch and energy envelope that dominate the entire utterance.

\textbf{Semantic coherence.} Table~\ref{tab:fig2_corr} (bottom) reveals acoustically meaningful patterns: ``loud'' correlates most strongly with energy ($r = +0.64$), confirming the model's attention peaks where audio is loudest. ``Nasal'' shows highest energy correlation ($r = +0.67$), consistent with increased spectral energy. Words describing emotional arousal (``nervous'': $r = +0.47$; ``dramatic'': $r = +0.46$) correlate more with energy than F0, while ``confident'' shows the opposite ($r_\text{F0} = +0.40$), consistent with pitch-raising in confident speech. These patterns demonstrate cross-attention functionally grounds style semantics in acoustic reality.

\subsection{Experiment 3: Layer and step dynamics}
\label{sec:fig3}

We decompose attention importance by transformer layer and ODE step separately for each token category, and also track attention entropy to measure selectivity.

\vspace{-7.5mm}
\begin{table}[h]
  \caption{Pearson $r$ between attention and acoustic features. Top: by category. Bottom: selected style words ($n \!\geq\! 224$).}
  \label{tab:fig2_corr}
  \centering
  \small
  \begin{tabular}{@{}lrrc@{}}
    \toprule
    & $\bar{r}_\text{F0}$ & $\bar{r}_\text{Energy}$ & $n$ \\
    \midrule
    \multicolumn{4}{@{}l}{\textit{By category}} \\
    {\color{csty}Style}   & $+0.21$ & $+0.28$ &  7{,}968 \\
    {\color{ccon}Content} & $+0.50$ & $+0.54$ &  8{,}480 \\
    {\color{cfun}Function}& $+0.11$ & $+0.09$ & 38{,}432 \\
    \midrule
    \multicolumn{4}{@{}l}{\textit{Selected style words}} \\
    loud       & $+0.49$ & $+0.64$ & 256 \\
    nasal      & $+0.41$ & $+0.67$ & 288 \\
    confident  & $+0.40$ & $+0.30$ & 256 \\
    nervous    & $+0.37$ & $+0.47$ & 352 \\
    robotic    & $+0.32$ & $+0.56$ & 384 \\
    dramatic   & $+0.30$ & $+0.46$ & 544 \\
    calm       & $+0.27$ & $+0.40$ & 224 \\
    \bottomrule
  \end{tabular}
\end{table}

\begin{figure}[h]
  \centering
  \includegraphics[width=\linewidth]{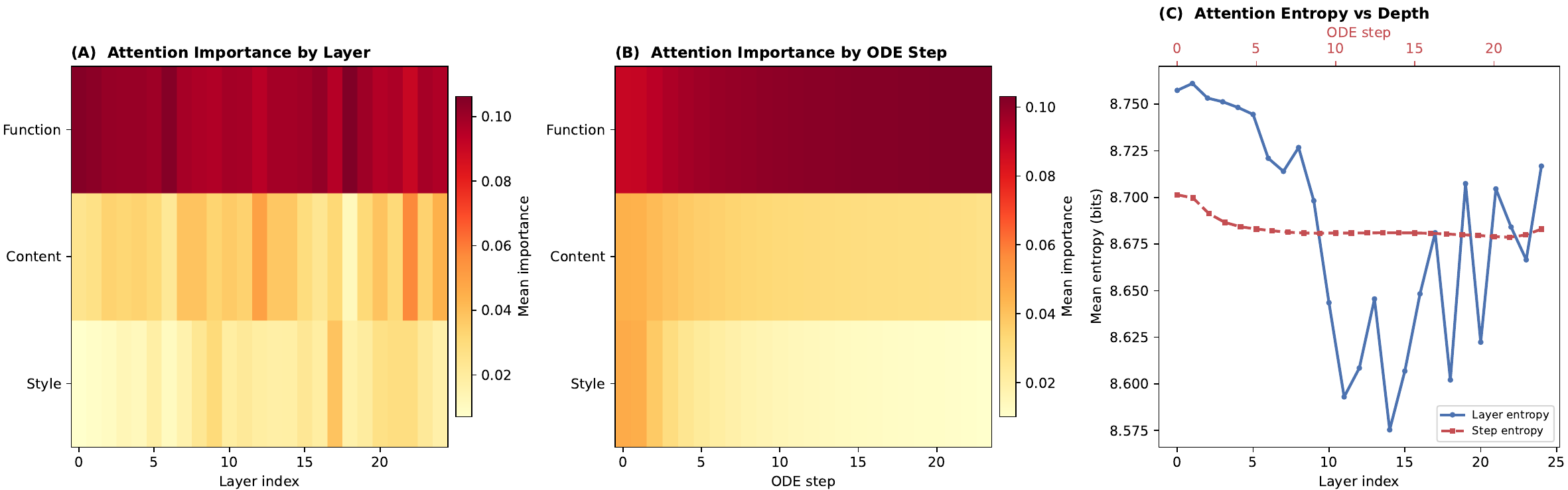}
  \caption{Layer and step dynamics. (A)~$I^{({\color{cpurple}l})}_\mathcal{C}$ by {\color{cpurple}layer}. (B)~$I^{({\color{cteal}s})}_\mathcal{C}$ by {\color{cteal}ODE step}. (C)~Entropy vs.\ {\color{cpurple}layer} (solid) and {\color{cteal}step} (dashed).}
  \label{fig:figure3}
\end{figure}

\textbf{Layer dynamics} (Figure~\ref{fig:figure3}A, Table~\ref{tab:layer_step}): Style-token importance increases monotonically from early to late layers, with peak importance at layer~17 ($I^{(17)}_\text{sty} = 0.034$). The late-to-early ratio is $R_\text{sty} = 1.28$, meaning the average style importance in layers 13--24 is 28\% higher than in layers 0--12. Content tokens peak even later at layer~22 ($I^{(22)}_\text{con} = 0.061$, $R_\text{con} = 1.07$), suggesting the network resolves style modulation first and then refines speaker identity in the deepest layers. Function tokens, by contrast, show flat or slightly declining importance across depth ($R_\text{fn} = 0.98$), meaning deeper layers \emph{selectively amplify} semantically meaningful tokens while suppressing grammatical scaffolding. This layer-wise hierarchy is reminiscent of findings in vision transformers~\cite{chefer2021generic,dosovitskiy2021vit}, where deeper layers produce more semantically refined attention patterns, and aligns with hierarchical feature learning observed in diffusion models~\cite{ho2020ddpm,peebles2023dit}.

\textbf{Step dynamics} (Figure~\ref{fig:figure3}B): Style importance peaks at ODE step $s\!=\!0$ ($I_\text{sty}^{(0)} = 0.053$) and declines to $0.010$ by $s = 23$, a $5.2\times$ decay, the largest among all categories. Content tokens decay more gradually ($1.7\times$: $0.048 \to 0.029$), suggesting speaker-identity features remain relevant throughout denoising. Most strikingly, function tokens show the \emph{opposite} trend: their importance \emph{increases} from $0.086$ at step~0 to $0.103$ at step~23 ($D_\text{fn} = 0.84\times$, i.e., rising). This crossover reveals that in early steps, the network primarily attends to style and content tokens to establish the global acoustic scaffold, while in later steps it shifts attention toward function tokens for fine-grained sequential structure such as phrasing and timing. This mirrors the coarse-to-fine dynamics observed in image diffusion~\cite{hertz2023p2p}.

\textbf{Entropy dynamics} (Figure~\ref{fig:figure3}C): Layer entropy ranges from 8.54 to 8.76 bits. The minimum occurs at layer~18 ($H^{(18)} = 8.54$~bits), directly adjacent to the style importance peak at layer~17. This co-occurrence is not coincidental: it indicates the network becomes \emph{maximally selective}, concentrating attention on fewer but more relevant tokens, at precisely the layers most critical for style conditioning. In contrast, step entropy is nearly constant ($\Delta H < 0.03$~bits across all 24 steps), indicating that the \emph{breadth} of attention remains stable throughout denoising even as its \emph{distribution} across token categories shifts dramatically.

\vspace{-7.5mm}
\begin{table}[h]
  \caption{Layer and step dynamics summary. $R$: late-to-early importance ratio (layers 13--24 vs.\ 0--12). $D$: first-to-last step decay ratio. Peak: layer or step of maximum importance.}
  \label{tab:layer_step}
  \centering
  \small
  \resizebox{\columnwidth}{!}{%
  \begin{tabular}{@{}lccc|ccc@{}}
    \toprule
    & \multicolumn{3}{c|}{\textbf{Layer dynamics}} & \multicolumn{3}{c}{\textbf{Step dynamics}} \\
    \textbf{Cat.} & Peak $l$ & $I^{(l)}_\text{peak}$ & $R$ & Peak $s$ & $D$ & Trend \\
    \midrule
    {\color{csty}Style}    & 17 & 0.034 & 1.28 & 0 & 5.2$\times$ & $\searrow$ \\
    {\color{ccon}Content}  & 22 & 0.061 & 1.07 & 0 & 1.7$\times$ & $\searrow$ \\
    {\color{cfun}Function} & 18 & 0.108 & 0.98 & 23 & 0.84$\times$ & $\nearrow$ \\
    \bottomrule
  \end{tabular}%
  }
\end{table}

\section{Discussion and Conclusion}
\label{sec:discussion}

Our three experiments converge on a unified picture: style-captioned TTS implements cross-attention as a \emph{hierarchical global conditioning channel}. Style tokens distribute attention uniformly across time ($9.2\times$ lower variance than function tokens, $d = {-}1.16$), correlate with acoustic features in semantically coherent patterns (``loud'' $\leftrightarrow$ energy, $r = +0.64$), and follow a coarse-to-fine schedule where early ODE steps establish global structure and deep transformer layers progressively refine acoustic details. This constitutes the first quantitative evidence that cross-attention in flow-matching TTS functions as a global modulation mechanism, fundamentally distinct from the temporal alignment role it plays in autoregressive TTS~\cite{ren2021fastspeech2}.

The per-word analysis reveals a clear \emph{division of labor}: acoustically grounded words (``loud'', ``nasal'') partially localize to regions where their effect is strongest, while abstract descriptors (``cheerful'', ``deep'') remain uniformly distributed. Early ODE steps establish coarse global structure, deeper layers refine acoustic details, and attention narrows at critical layers (17--18) to focus on informative tokens---enabling both efficient global style conditioning and fine-grained sequential control within a single mechanism.

\textbf{Limitations.} Our analysis is limited to one model (CapSpeech) and synthetic prompts over 30 style words. Future work should extend to: (1)~other flow-matching and diffusion-based TTS architectures, (2)~naturally occurring prompts from user studies, (3)~causal intervention via attention editing~\cite{hertz2023p2p}, (4)~per-head analysis to identify specialized style-encoding heads, and (5)~comparison with baseline attention patterns to quantify learned structure.

\clearpage
\section{Use of Generative AI Disclosure}
In preparing this manuscript, the authors used generative AI tools for language refinement (rephrasing and improving the clarity of author-written text) and as a coding assistant (helping write and debug software for experiments and analysis). All research contributions, including the methodology, experimental design, results, and scientific claims, are the authors' own. The authors reviewed and verified all AI-assisted text and code, and take full responsibility for the content of this paper.
{\hbadness=10000
\bibliographystyle{IEEEtran}
\bibliography{refs}
}

\clearpage

\end{document}